\documentclass[10pt]{article}
\usepackage[margin=0.80in]{geometry}
\usepackage{times}
\usepackage{microtype}
\usepackage{amsmath,amssymb,amsthm,mathtools}
\usepackage{booktabs}
\usepackage{multirow}
\usepackage{graphicx}
\usepackage{subcaption}
\usepackage{enumitem}
\usepackage[numbers,sort&compress]{natbib}
\usepackage[colorlinks=true,linkcolor=blue,citecolor=blue,urlcolor=blue]{hyperref}
\usepackage[T1]{fontenc}
\usepackage{url}
\newtheorem{definition}{Definition}
\newtheorem{principle}{Principle}

\newtheorem{axiom}{Axiom}

\newcommand{\M}{\mathcal M}

\newcommand{\U}{\mathcal U}

\newcommand{\Zcal}{\mathcal Z}
\newcommand{\R}{\mathsf R}
\newcommand{\I}{\mathsf I}
\newcommand{\V}{\mathsf V}
\newcommand{\Mode}{\psi}
\newcommand{\Prob}{\mathbb P}
\newcommand{\Qlaw}{\mathbb Q}
\newcommand{\Hh}{\mathcal H}
\newcommand{\Kcap}{K}
\newcommand{\CC}{\operatorname{CC}}

\newcommand{\argmin}{\operatorname*{arg\,min}}

\title{Mirror Horizon: Viable Path Entropy as a Measure of Bounded Reflection}
\author{Tiantian (Crystal) Zhang\\Department of Computer Science, Columbia University\\\texttt{t.zhang8@columbia.edu}}
\date{}

\begin{document}
\maketitle

\begin{abstract}
Mirror Theory proposes that an intelligent system should be studied not only by what it represents, but by what coherent continuations it can sustain under repeated reflection. We make this claim operational through \emph{viable path entropy} (VPE), a finite-budget measure of verified continuation capacity. Given a mirror state, a rollout protocol, a verifier, and a mode map, VPE decomposes bounded capability into two parts: the probability of reaching a viable continuation and the diversity of verified continuation modes reached among successful rollouts. This paper restores the full theoretical scaffold behind the measure: intuition as local underdetermining constraint, taste as invariant-selecting pressure, reflection as taste-guided resolution of underdetermination, and geometry as the learned structure that makes future reflection stable. We then instantiate the theory in language-model reasoning experiments on GSM8K. Across Qwen2.5-Instruct models, 32 sampled rollouts per problem, and two reflection horizons, increasing the token budget from 96 to 160 substantially expands verified reachability, reduces zero-reachability, increases verified-mode entropy, and improves smoothed VPE. At 160 tokens, Qwen2.5-1.5B realizes the strongest mirror horizon among the tested models, even though Qwen2.5-3B has more parameters. This shows that mirror horizon is not parameter count, but accessible verified continuation capacity under a bounded reflection protocol. The result supports Mirror Theory as a measure-level account: capability is the structure of viable continuations made reachable, not merely one-shot accuracy or pass@k.
\end{abstract}

\section{Introduction}

Large language models are usually evaluated by loss, accuracy, pass@k, benchmark score, or reward. These quantities are important, but they collapse a system's internal capacity into a single outcome statistic. A model may solve a problem once, fail most other attempts, and nevertheless obtain a nonzero pass@k. Another model may solve fewer problems, but when it succeeds it may reveal many distinct verified solution modes. A third model may have high raw output entropy but little verified structure. These distinctions matter if we want to understand capability as more than single-answer success.

Mirror Theory begins from a different primitive. An intelligent system does not merely store a representation. It maintains an internal world that survives and unfolds through repeated reflection. In the earlier mathematical notes, this distinction was summarized as: \emph{a representation encodes; a mirror survives reflection}. The present paper turns that claim into a measurable object. A mirror is not identified only with a hidden vector, a prompt, a model checkpoint, or a set of beliefs. It is identified with the continuation law it induces: what futures become reachable from that state, which futures remain viable, and how many distinct verified modes those viable futures occupy.

The central proposal is \emph{viable path entropy}. Given finite budget $B$, finite horizon $T$, a continuation law, a verifier $\V$, and a mode map $\Mode$, we define
\begin{equation}
    \Hh_{B,T}(M)=\log \Pr[\V=1] + H(\Mode\mid \V=1),
\end{equation}
when the viability probability is nonzero. The first term is verified reachability. The second term is verified-mode diversity. Together they measure a finite-budget mirror horizon: the effective number of coherent semantic continuation modes reachable from the mirror.

The theoretical ambition is not to prove a universal monotone scaling law. In fact, our experiments show why that would be the wrong claim. A larger model need not have a larger measured mirror horizon under a fixed protocol. Parameter count, latent capability, and accessible verified continuation capacity are different objects. The present paper argues for the third object. A mirror's horizon is protocol-conditioned: it depends on the model, task, prompt, rollout budget, sampling rule, verifier, and semantic mode map.

Our empirical study is deliberately simple. We sample bounded rollouts from Qwen2.5-Instruct models on GSM8K, verify final numeric correctness, and group verified solutions into coarse reasoning modes. We run 30 GSM8K problems with 32 sampled rollouts per problem, temperature $0.8$, top-p $0.95$, and a heuristic mode map. Two horizons are compared: 96 and 160 maximum new tokens. The 96-token run includes Qwen2.5-0.5B and Qwen2.5-1.5B. The 160-token run includes Qwen2.5-0.5B, 1.5B, and 3B.

The main findings are straightforward. First, increasing reflection budget from 96 to 160 tokens expands mirror horizon for both 0.5B and 1.5B: verified probability rises, pass@32 rises, zero-verified fraction drops, verified-mode entropy rises, and smoothed VPE improves. Second, at 160 tokens Qwen2.5-1.5B dominates the tested models on every core component of the measure. Third, Qwen2.5-3B is not best: it has higher average verified probability than 0.5B but lower breadth across problems, lower verified-mode entropy than 1.5B, and lower overall VPE than 1.5B. This is not a contradiction of Mirror Theory. It is the point: mirror horizon measures accessible verified continuation capacity, not raw size.

\paragraph{Contributions.}
We make four contributions.
\begin{enumerate}[leftmargin=1.5em]
    \item We give a formal path-space version of Mirror Theory in which mirrors are continuation laws rather than static representations.
    \item We define viable path entropy as a finite-budget mirror horizon, decomposing bounded capability into verified reachability and verified-mode diversity.
    \item We connect older theoretical components -- intuition, taste, geometry, constructive compatibility, and invariant survival -- to the measurable VPE object.
    \item We report real GSM8K rollout experiments showing that reflection budget expands verified continuation capacity, and we add empirical coverage curves showing how mirror horizon unfolds as rollout budget increases.
\end{enumerate}

\section{Mirror Theory: the formal scaffold}

This section restores the theory that motivates the empirical measure. The goal is not to introduce a new state-transition formalism for its own sake. The goal is to define what kind of internal object can be evaluated by viable continuation capacity.

\subsection{Reflection, intuition, taste, and geometry}

Let there be an external structured world $W$. The system does not access $W$ directly. It receives finite evidence $e_t$, and this evidence gives only local constraints. It does not determine a full internal world. A current mirror $M_t$ is updated through reflection, but reflection is not merely a transition function. It is a selection among admissible internal continuations.

\begin{definition}[Mirror geometry]
At time $t$, a mirror geometry is
\begin{equation}
    G_t=(\M_t,d_t,\U_t,\mathcal I_t),
\end{equation}
where $\M_t$ is the current space of possible mirrors, $d_t$ is a learned notion of nearness or deformation cost, $\U_t$ is an admissibility generator, and $\mathcal I_t$ is a family of valued invariants. The geometry specifies which mirrors are near, which updates are admissible, and which internal properties must survive reflection.
\end{definition}

\begin{definition}[Intuition as local constraint]
Intuition is not a full next mirror. It is the local constraint imposed by evidence on possible mirrors. Given geometry $G_t$, evidence $e_t$ induces an incompatibility functional
\begin{equation}
    \I_{G_t}(e_t):\M_t\to \mathbb R_{\ge 0}.
\end{equation}
Smaller $\I_{G_t}(e_t)(M')$ means that $M'$ is more compatible with the current evidence. Intuition is local, world-induced, and underdetermining.
\end{definition}

\begin{definition}[Taste as invariant-selecting pressure]
Taste is the selection structure that ranks admissible mirror continuations by what should survive. It may be a preorder $\preceq_{\tau,G_t}$ over $\M_t$, or, when scalarizable, a potential $\tau_{G_t}:\M_t\to\mathbb R$. Conceptually, taste is not merely preference over next states. It is pressure toward continuations that preserve valued invariants: coherence, fidelity, identity, usefulness, simplicity, compressibility, actionability, or task success.
\end{definition}

Given current mirror $M_t$ and evidence $e_t$, intuition first induces an admissible candidate set
\begin{equation}
    \U_{G_t}(M_t,e_t)=\{M'\in \M_t:\I_{G_t}(e_t)(M')\le \epsilon_t,\ d_t(M_t,M')\le \rho_t\}.
\end{equation}
Taste then selects a continuation:
\begin{equation}
    M_{t+1}\in \max_{\preceq_{\tau,G_t}} \U_{G_t}(M_t,e_t).
\end{equation}
Repeated intuition-taste interaction is reflection. The mirror is not simply $M_t$ as a point; it is the persistent internal structure $(M_t,G_t)$ shaped by such reflections.

\begin{principle}[Why reflection learns geometry]
Finite evidence gives only local constraints. Taste selects which compatible continuations are worth preserving. Geometry is the learned structure that makes such selections stable across time. In short: intuition gives local contact; taste decides what survives; geometry makes survival stable.
\end{principle}

This statement is the conceptual core of Mirror Theory. It also explains why a purely one-step rationalization result is not enough. Any single deterministic update can be rationalized after the fact by a preorder. What matters diachronically is whether repeated reflection learns a stable geometry of admissibility, nearness, and invariant survival.

\subsection{Optimization form}

The same theory can be written as constrained or proximal optimization. Let
\begin{equation}
    C_t(M')=\I_{G_t}(e_t)(M')
\end{equation}
be intuition incompatibility, and let $d_t(M_t,M')^2$ be a geometry-dependent movement cost. Then reflection can be represented as
\begin{equation}
M_{t+1}\in \argmin_{M'\in\M_t}
\left[C_t(M')+\lambda d_t(M_t,M')^2-\beta\tau_{G_t}(M')+\gamma\operatorname{Rupt}(M_t,M')\right].
\label{eq:proximal_reflection}
\end{equation}
This expression is not meant to reduce Mirror Theory to standard optimization. It clarifies the roles of the components: evidence fit, geometric continuity, taste value, and invariant survival. The geometry $G_t$ itself is learned because the same optimization must be solved repeatedly under underdetermination.

\subsection{Constructive compatibility}

A possibility $p$ is an encounter or intervention: a continuation, prompt edit, retrieved document, theorem, action, candidate solution, or environmental event. It induces a counterfactual mirror
\begin{equation}
    M^{(p)}=\R(M,\I(p)).
\end{equation}
The older constructive compatibility note expressed the desired regime as: not same, not random, but absorbably expansive. Similarity alone leads to stagnation. Novelty alone can rupture invariants. Constructive compatibility means the possibility expands the mirror while preserving what must survive.

In the earlier geometric version, constructive compatibility was written schematically as
\begin{equation}
    \CC(M,p)=\operatorname{Exp}(M,p)-\lambda \operatorname{Rupt}(M,p)-\mu \operatorname{Red}(M,p),
\end{equation}
where expansion measures how far the mirror moves, rupture penalizes invariant failure, and redundancy penalizes pure sameness. The VPE formulation makes this operational:
\begin{equation}
    \CC_{B,T}(M,p)=\Hh_{B,T}(M^{(p)})-\Hh_{B,T}(M).
\label{eq:cc_vpe}
\end{equation}
A possibility is constructively compatible if it increases finite-budget viable continuation capacity. In this sense, VPE is the measurable version of absorbable expansion.

\section{Viable path entropy}

We now define the measure used in the experiments. The definitions are stated for an abstract mirror $M$ but instantiated later by an LLM, a prompt, a rollout procedure, a numeric verifier, and a solution-mode map.

\subsection{Path-space setup}

\begin{definition}[Budget and horizon]
Let $B$ denote a finite resource budget: rollout count, compute, memory, token length, intervention size, or time. Let $T$ denote a finite continuation horizon. All mirror quantities are indexed by $(B,T)$.
\end{definition}

\begin{definition}[Continuation path space]
For a mirror state $M$, define the $T$-step continuation path space
\begin{equation}
    \Gamma_T(M)=\{\gamma=(M_0,M_1,\ldots,M_T):M_0=M\}.
\end{equation}
A system with budget $B$ induces a probability law $\Prob^M_{B,T}$ over $\Gamma_T(M)$. In an LLM, this law is induced by a prompt, decoding rule, sampling temperature, token budget, and model parameters.
\end{definition}

\begin{definition}[Viability verifier]
A viability verifier is a measurable map
\begin{equation}
    \V:\Gamma_T(M)\to\{0,1\},
\end{equation}
or a soft version $\V:\Gamma_T(M)\to[0,1]$. A continuation is viable when it satisfies the task's coherence requirements: correctness, semantic consistency, factuality, safety, invariant preservation, or another domain-specific criterion.
\end{definition}

\begin{definition}[Mode map]
A mode map is a measurable function
\begin{equation}
    \Mode:\Gamma_T(M)\to \Zcal
\end{equation}
that maps a continuation to a semantic outcome mode. The mode map prevents random token diversity from being counted as meaningful continuation capacity. In reasoning tasks, modes can be coarse solution strategies, operation signatures, proof forms, or clusters of verified rationales.
\end{definition}

\subsection{Viable continuation capacity}

\begin{definition}[Viable continuation capacity]
For discrete modes, define
\begin{equation}
    \Kcap_{B,T}(M)=\Pr_{\gamma\sim \Prob^M_{B,T}}[\V(\gamma)=1]\,\exp\left(H(\Mode(\gamma)\mid \V(\gamma)=1)\right).
\label{eq:kcap}
\end{equation}
If $\Pr[\V=1]=0$, set $\Kcap_{B,T}(M)=0$.
\end{definition}

Thus
\begin{equation}
    \Kcap_{B,T}(M)=\text{viability probability}\times \text{effective number of viable semantic modes}.
\end{equation}

\begin{definition}[Viable path entropy]
The viable path entropy of $M$, also called its mirror horizon, is
\begin{equation}
    \Hh_{B,T}(M)=\log \Kcap_{B,T}(M).
\end{equation}
When $\Pr[\V=1]>0$,
\begin{equation}
    \Hh_{B,T}(M)=\log \Pr[\V=1]+H(\Mode\mid \V=1).
\label{eq:vpe}
\end{equation}
If $\Pr[\V=1]=0$, set $\Hh_{B,T}(M)=-\infty$.
\end{definition}

The two terms of Equation~\ref{eq:vpe} should not be hidden behind a single scalar. The first is reachability: whether verified continuations are accessible at all. The second is diversity: how many distinct verified modes are supported once viability is reached. The scalar is useful, but the decomposition is the main empirical object.

\subsection{Mirror as continuation law}

\begin{definition}[Mirror]
A mirror is an internal state considered through the viable continuation law it induces:
\begin{equation}
    M\longmapsto \Prob^M_{B,T}.
\end{equation}
Its strength is $\Hh_{B,T}(M)$. Two states are mirror-equivalent at $(B,T,\V,\Mode)$ if they induce the same viable-mode law. Let $\Qlaw^M_{B,T}$ be the distribution on $\Zcal\cup\{\bot\}$ defined by
\begin{equation}
    \Qlaw^M_{B,T}(\bot)=1-\Pr[\V=1],\qquad
    \Qlaw^M_{B,T}(z)=\Pr[\V=1,\Mode=z].
\end{equation}
Then $M\sim N$ if $\Qlaw^M_{B,T}=\Qlaw^N_{B,T}$.
\end{definition}

A mirror is therefore not defined by what it contains, but by what viable continuations it makes reachable. This is the technical version of the conceptual statement: a representation encodes; a mirror survives reflection.

\subsection{Estimator}

Given rollouts $\gamma_1,\ldots,\gamma_N\sim \Prob^M_{B,T}$, estimate
\begin{equation}
    \hat p_M=\frac{1}{N}\sum_i \V(\gamma_i).
\end{equation}
Among viable rollouts, estimate the mode distribution $\hat\pi_M(z)$. Then
\begin{equation}
    \widehat{\Hh}_{B,T}(M)=\log(\hat p_M+\epsilon)-\sum_z \hat\pi_M(z)\log\hat\pi_M(z).
\end{equation}
For finite samples we also report a smoothed version using
\begin{equation}
    \hat p_{\mathrm{smooth}}=\frac{n_{\mathrm{verified}}+0.5}{n_{\mathrm{rollouts}}+1}.
\end{equation}
Raw VPE is theory-faithful and harshly penalizes zero-reachability. Smoothed VPE is more stable for small rollout budgets.

We also estimate a non-parametric 	extit{coverage horizon} from the same rollouts. For a budget of $b$ rollouts, define
\begin{equation}
    C_b(M)=\left|\{\Mode(\gamma_i):\V(\gamma_i)=1,\ i=1,\ldots,b\}\right|,
\end{equation}
the number of distinct verified modes reached by those $b$ rollouts. The empirical coverage capacity is
\begin{equation}
    K_b^{\mathrm{cov}}(M)=\mathbb E[C_b(M)],\qquad
    \Hh_b^{\mathrm{cov}}(M)=\log(1+K_b^{\mathrm{cov}}(M)).
\label{eq:coverage_horizon}
\end{equation}
In the experiments, $K_b^{\mathrm{cov}}$ is estimated by uniform subsampling from the 32 observed rollouts per problem for $b\in\{1,2,4,8,16,32\}$. This estimator makes no iid or exponential-hit assumption; it directly asks how many distinct verified modes are reached by a bounded reflection budget.

\section{Related work}

\paragraph{Coverage and test-time extraction.}
The Coverage Principle argues that pre-training enables post-training and Best-of-N style methods by placing sufficient probability mass on high-quality responses, and that such coverage can be more predictive of downstream success than cross-entropy in relevant regimes \citep{chen2025coverage}. Our first term, $\Pr[\V=1]$, is a rollout-side coverage quantity. Our contribution is to refine coverage by asking what structure exists inside the verified region: how many distinct verified modes are reachable.

\paragraph{Reasoning and verifier-guided sampling.}
Chain-of-thought prompting and self-consistency show that multiple samples can expose reasoning capability not visible in one-shot decoding \citep{wei2022cot,wang2023selfconsistency}. GSM8K provides a verifier-backed setting for evaluating math reasoning \citep{cobbe2021gsm8k}. VPE uses the same sampling-and-verification backbone but adds a mode map and entropy over verified modes.

\paragraph{Scaling laws and frontiers.}
Classical scaling laws study loss as a function of model size, data, and compute \citep{kaplan2020scaling,hoffmann2022chinchilla}. Effective Frontier approaches interpret scaling as the movement of a resource-dependent frontier through long-tailed pattern space \citep{zou2026effective}. Our measure is not a scaling law in parameter count; it is a protocol-conditioned horizon. It can be used to study frontiers over rollout budget, token horizon, or verifier accessibility.

\paragraph{Epiplexity and bounded structure.}
Epiplexity aims to measure structural information learnable by computationally bounded observers, distinguishing useful structure from random time-bounded entropy \citep{finzi2026epiplexity}. Mirror horizon is complementary: it measures usable structure accessible during bounded rollout rather than learnable structure in a dataset. Epiplexity is therefore related work, not a foundation.

\paragraph{Reasoning compression and post-training.}
Recent work on compressed reasoning data studies how explicit, composed, and implicit chain-of-thought traces affect SFT and RLVR \citep{matsutani2026zipping}. This is relevant because compression changes the visibility and accessibility of continuation modes. VPE could measure how much verified-mode coverage remains after such compression.

\paragraph{Taste, rationalizability, and invariant survival.}
The selection spine of Mirror Theory connects to revealed preference and inverse reinforcement learning \citep{richards1966revealed,afriat1967utility,debreu1954representation}. The survival spine connects to inductive invariants and property testing \citep{garg2014ice,daskalakis2018testing}. The present paper does not claim novelty on either spine alone. Its claim is that VPE couples viability and mode structure into one operational mirror horizon.

\section{Experimental setup}

We instantiate mirror horizon in language-model math reasoning. A problem prompt defines the evidence. The model plus decoding procedure defines the rollout law. Each generated solution is a continuation path. The verifier checks final numeric correctness. A heuristic mode map groups verified solutions by coarse reasoning form.

\subsection{Task and models}

We use GSM8K test problems \citep{cobbe2021gsm8k}. The 96-token run uses Qwen2.5-0.5B-Instruct and Qwen2.5-1.5B-Instruct. The 160-token run uses Qwen2.5-0.5B-Instruct, Qwen2.5-1.5B-Instruct, and Qwen2.5-3B-Instruct. Both experiments use 30 randomly selected GSM8K test problems, 32 sampled rollouts per problem, temperature $0.8$, top-p $0.95$, and a heuristic mode map. The 96-token run uses maximum 96 new tokens; the 160-token run uses maximum 160 new tokens.

\begin{table}[t]
\centering
\caption{Experimental configurations. Both runs use GSM8K, 30 problems, 32 rollouts per problem, temperature $0.8$, top-p $0.95$, heuristic mode map, and Qwen2.5-Instruct models.}
\label{tab:setup}
\begin{tabular}{llllll}
\toprule
Run & Models & Problems & Rollouts & Horizon & Mode map \\
\midrule
96-token & 0.5B, 1.5B & 30 & 32 & 96 tokens & heuristic \\
160-token & 0.5B, 1.5B, 3B & 30 & 32 & 160 tokens & heuristic \\
\bottomrule
\end{tabular}
\end{table}

\subsection{Verifier}

For each rollout, the model is prompted to solve a GSM8K problem and end with a final numeric answer. We extract the predicted number using a regex that prioritizes answer-marker patterns such as \texttt{Answer: <number>} and then falls back to the last number in the generation. The verifier returns $1$ if the normalized predicted number equals the GSM8K gold answer and $0$ otherwise. This verifier is simple and strict. It may undercount valid reasoning with unusual formatting, but it gives a reproducible first operationalization.

\subsection{Mode map}

The heuristic mode map is deliberately transparent. For verified rollouts, it records: (i) a length bin (short, medium, long), (ii) an operator signature inferred from symbols and words such as addition, subtraction, multiplication, and division, (iii) an equation-count bin, and (iv) a reasoning-style indicator such as step-like versus direct. Invalid rollouts are assigned to a single invalid mode and are not counted in verified-mode entropy.

This mode map is not intended as the final semantic geometry. Its role is to demonstrate that VPE can distinguish verified reachability from verified-mode diversity. Future versions should test TF-IDF clusters, embedding clusters, symbolic operation templates, and human-audited reasoning modes.

\subsection{Metrics}

For each model and problem, we compute:
\begin{itemize}[leftmargin=1.5em]
    \item mean verified probability $\hat p$ across 32 rollouts;
    \item pass@32, equal to one if at least one rollout is verified;
    \item zero-verified indicator, equal to one if no rollout is verified;
    \item verified-mode entropy $H(\Mode\mid \V=1)$;
    \item raw VPE $\log(\hat p+\epsilon)+H(\Mode\mid\V=1)$;
    \item smoothed VPE $\log((n_{\mathrm{verified}}+0.5)/(n_{\mathrm{rollouts}}+1))+H(\Mode\mid\V=1)$;
    \item average verified rollouts and average verified modes per problem;
    \item empirical coverage curves $b\mapsto \Hh_b^{\mathrm{cov}}$ for $b\in\{1,2,4,8,16,32\}$.
\end{itemize}

The main text emphasizes smoothed VPE because it is more stable at 32 rollouts. Raw VPE is still reported because it faithfully penalizes zero-reachability.

\section{Results}

\subsection{Reflection budget expands mirror horizon}

Table~\ref{tab:96} shows the 96-token result. Qwen2.5-1.5B improves over Qwen2.5-0.5B on every component: verified probability, pass@32, zero-verified fraction, verified-mode entropy, raw VPE, smoothed VPE, verified rollouts, and verified modes.

\begin{table}[t]
\centering
\small
\caption{96-token GSM8K result. Values are averaged over 30 problems with 32 rollouts each.}
\label{tab:96}
\begin{tabular}{lrrrrrrrr}
\toprule
Model & $\Pr[V]$ & pass@32 & zero-frac & mode H & raw VPE & smooth VPE & verified & modes \\
\midrule
Qwen2.5-0.5B & 0.057 & 0.400 & 0.600 & 0.224 & -17.411 & -3.249 & 1.83 & 1.10 \\
Qwen2.5-1.5B & 0.097 & 0.567 & 0.433 & 0.507 & -12.687 & -2.450 & 3.10 & 1.90 \\
\bottomrule
\end{tabular}
\end{table}

Table~\ref{tab:160} shows the 160-token result. Increasing the reflection horizon makes the mirror-horizon signal much clearer. For 0.5B and 1.5B, both reachability and verified-mode diversity improve sharply. At 160 tokens, Qwen2.5-1.5B is strongest on every core metric.

\begin{table}[t]
\centering
\small
\caption{160-token GSM8K result. Values are averaged over 30 problems with 32 rollouts each.}
\label{tab:160}
\begin{tabular}{lrrrrrrrr}
\toprule
Model & $\Pr[V]$ & pass@32 & zero-frac & mode H & raw VPE & smooth VPE & verified & modes \\
\midrule
Qwen2.5-0.5B & 0.101 & 0.667 & 0.333 & 0.518 & -10.278 & -2.347 & 3.23 & 1.90 \\
Qwen2.5-1.5B & 0.257 & 0.867 & 0.133 & 0.894 & -4.243 & -1.038 & 8.23 & 3.57 \\
Qwen2.5-3B   & 0.227 & 0.633 & 0.367 & 0.560 & -10.617 & -1.948 & 7.27 & 2.20 \\
\bottomrule
\end{tabular}
\end{table}

Table~\ref{tab:delta} isolates the effect of increasing the reflection budget from 96 to 160 tokens for the two models present in both runs. The gain is large. For Qwen2.5-1.5B, verified probability increases by 0.160, pass@32 increases by 0.300, zero-verified fraction drops by 0.300, verified-mode entropy rises by 0.387, and smoothed VPE improves by 1.412 nats.

\begin{table}[t]
\centering
\small
\caption{Effect of increasing the reflection horizon from 96 to 160 tokens. Positive changes in $\Pr[V]$, pass@32, mode entropy, and VPE are improvements; negative change in zero-frac is an improvement.}
\label{tab:delta}
\begin{tabular}{lrrrrr}
\toprule
Model & $\Delta\Pr[V]$ & $\Delta$pass@32 & $\Delta$zero-frac & $\Delta$mode H & $\Delta$smooth VPE \\
\midrule
Qwen2.5-0.5B & +0.044 & +0.267 & -0.267 & +0.294 & +0.902 \\
Qwen2.5-1.5B & +0.160 & +0.300 & -0.300 & +0.387 & +1.412 \\
\bottomrule
\end{tabular}
\end{table}

\begin{figure}[t]
\centering
\includegraphics[width=0.98\linewidth]{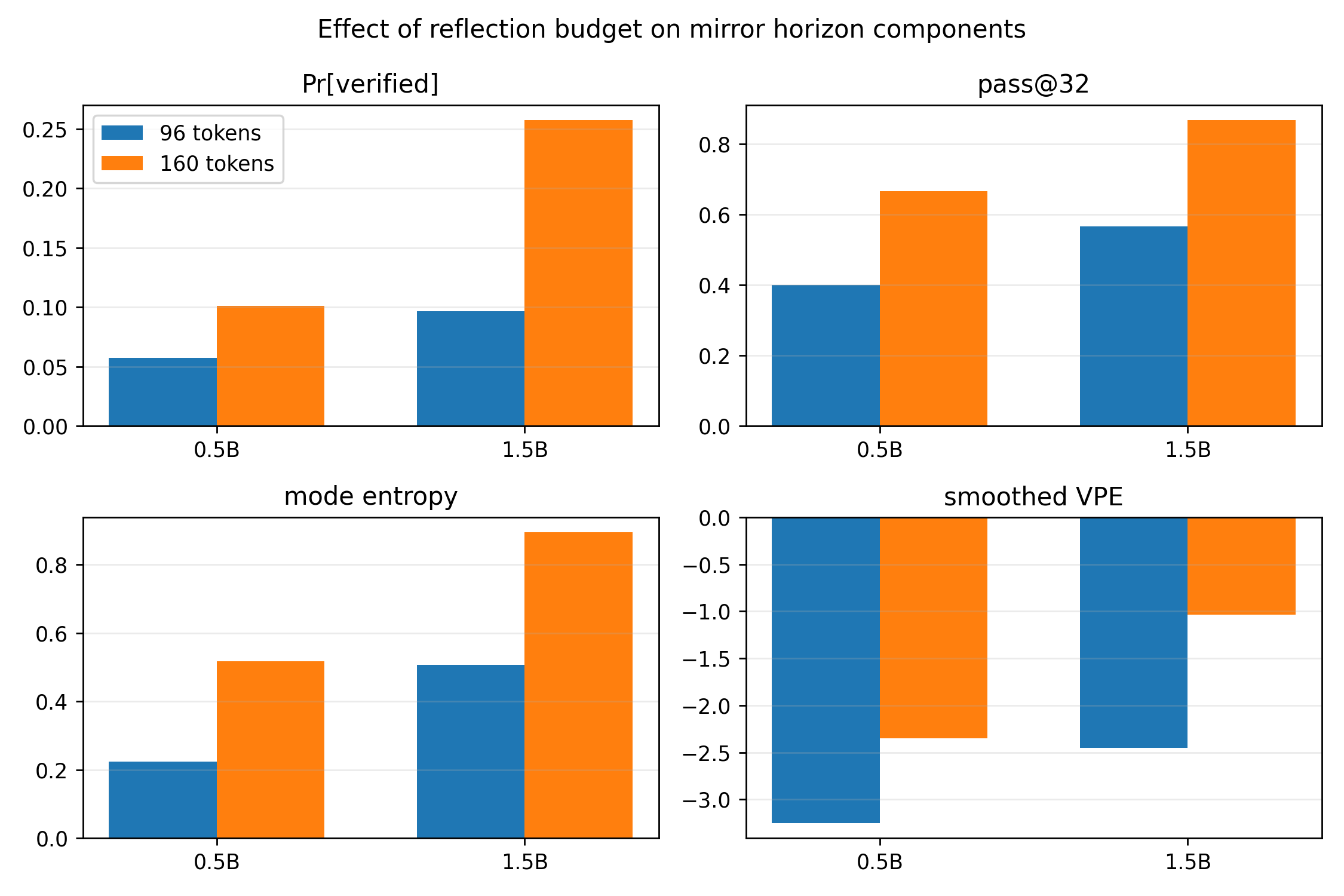}
\caption{Increasing the reflection horizon from 96 to 160 tokens expands mirror horizon components. Both 0.5B and 1.5B improve in verified probability, pass@32, verified-mode entropy, and smoothed VPE.}
\label{fig:budget}
\end{figure}

\subsection{Empirical coverage curves: mirror horizon unfolds with rollout budget}

The previous tables report one horizon value at the full 32-rollout budget. Mirror Theory, however, predicts a curve: as the reflection budget increases, more verified continuations should become reachable and more verified modes should be uncovered. We therefore reuse the same 160-token rollout table and estimate $\Hh_b^{\mathrm{cov}}$ for $b\in\{1,2,4,8,16,32\}$ by subsampling from the 32 observed rollouts. This requires no additional model inference.

Table~\ref{tab:coverage32} shows the endpoint at $b=32$. Qwen2.5-1.5B reaches an expected 3.567 distinct verified modes per problem, compared with 1.900 for 0.5B and 2.200 for 3B. Its empirical coverage horizon is therefore largest, and it also has the highest pass@32 and lowest zero-coverage fraction.

\begin{table}[t]
\centering
\small
\caption{Empirical verified-mode coverage at the full 32-rollout budget. $K_{32}$ is the expected number of distinct verified modes reached; $\Hh_{32}^{\mathrm{cov}}=\log(1+K_{32})$.}
\label{tab:coverage32}
\begin{tabular}{lrrrrr}
\toprule
Model & $K_{32}$ & $\Hh_{32}^{\mathrm{cov}}$ & pass@32 & zero-cov & cond. modes \\
\midrule
Qwen2.5-0.5B & 1.900 & 0.824 & 0.667 & 0.333 & 1.900 \\
Qwen2.5-1.5B & 3.567 & 1.292 & 0.867 & 0.133 & 3.567 \\
Qwen2.5-3B   & 2.200 & 0.880 & 0.633 & 0.367 & 2.200 \\
\bottomrule
\end{tabular}
\end{table}

Table~\ref{tab:coverage_curve_values} reports the coverage horizon across budgets. The 1.5B model dominates the curve at every budget, not only at the endpoint. This is the cleanest empirical expression of the revised definition: a stronger mirror is the one whose bounded reflections reach more distinct verified modes.

\begin{table}[t]
\centering
\small
\caption{Coverage-horizon curve values $\Hh_b^{\mathrm{cov}}=\log(1+K_b)$ estimated by subsampling observed rollouts. Higher is better.}
\label{tab:coverage_curve_values}
\begin{tabular}{rrrr}
\toprule
Budget $b$ & Qwen2.5-0.5B & Qwen2.5-1.5B & Qwen2.5-3B \\
\midrule
1  & 0.088 & 0.209 & 0.177 \\
2  & 0.154 & 0.346 & 0.281 \\
4  & 0.255 & 0.533 & 0.412 \\
8  & 0.399 & 0.767 & 0.557 \\
16 & 0.590 & 1.031 & 0.713 \\
32 & 0.824 & 1.292 & 0.880 \\
\bottomrule
\end{tabular}
\end{table}

\begin{figure}[t]
\centering
\includegraphics[width=0.98\linewidth]{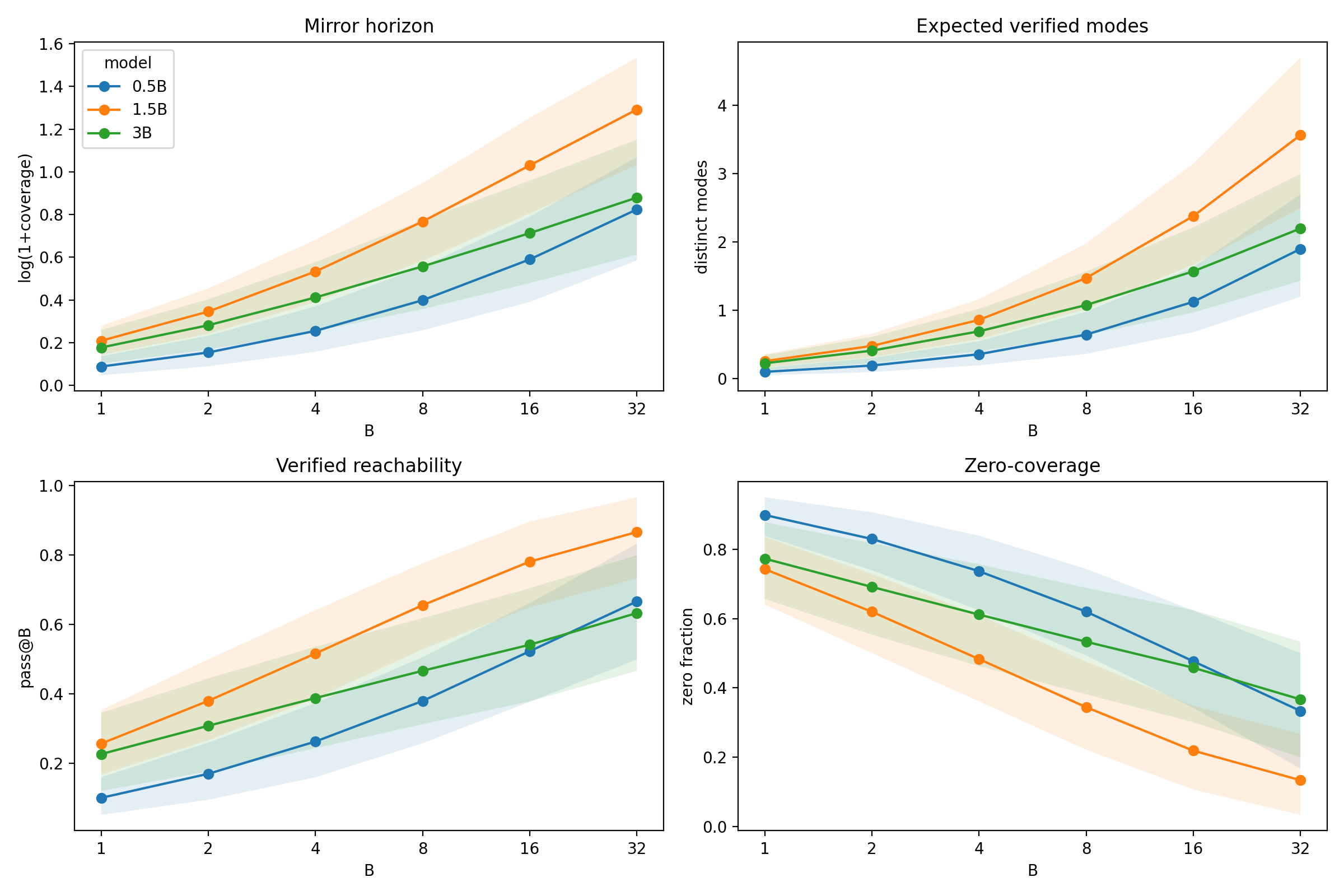}
\caption{Empirical coverage curves from the 160-token rollout table. The top-left panel is the main mirror-horizon curve: $\log(1+\mathbb E[\#\text{ distinct verified modes reached}])$ as rollout budget increases. The remaining panels show the decomposition into expected verified modes, pass@b, and zero-coverage probability. Qwen2.5-1.5B dominates the coverage curve across budgets.}
\label{fig:coverage_dashboard}
\end{figure}

This result is stronger than a single VPE-vs-scale plot because it measures how horizon unfolds with bounded reflection. It also avoids imposing a closed-form hitting probability such as $1-(1-p)^b$. The expectation is empirical: given the rollouts actually observed, how many verified modes become reachable as budget grows?

\subsection{At fixed budget, VPE separates breadth from depth}

At 160 tokens, Qwen2.5-1.5B is best overall. It has highest verified probability, highest pass@32, lowest zero-verified fraction, highest verified-mode entropy, and highest smoothed VPE. Qwen2.5-3B is more subtle. It has higher average verified probability than 0.5B (0.227 versus 0.101), meaning it produces more verified rollouts on average. But it has lower pass@32 than 0.5B (0.633 versus 0.667), meaning its successes are concentrated on fewer problems. It also has much lower verified-mode entropy than 1.5B.

This is exactly the kind of distinction VPE is meant to reveal. Pass@32 only captures whether at least one verified continuation appears. Mean verified probability captures average rollout correctness. Mode entropy captures diversity among successful continuations. VPE combines reachability and mode diversity, and the decomposition explains the model behavior more clearly than a single benchmark number.

\begin{figure}[t]
\centering
\begin{subfigure}{0.32\linewidth}
\includegraphics[width=\linewidth]{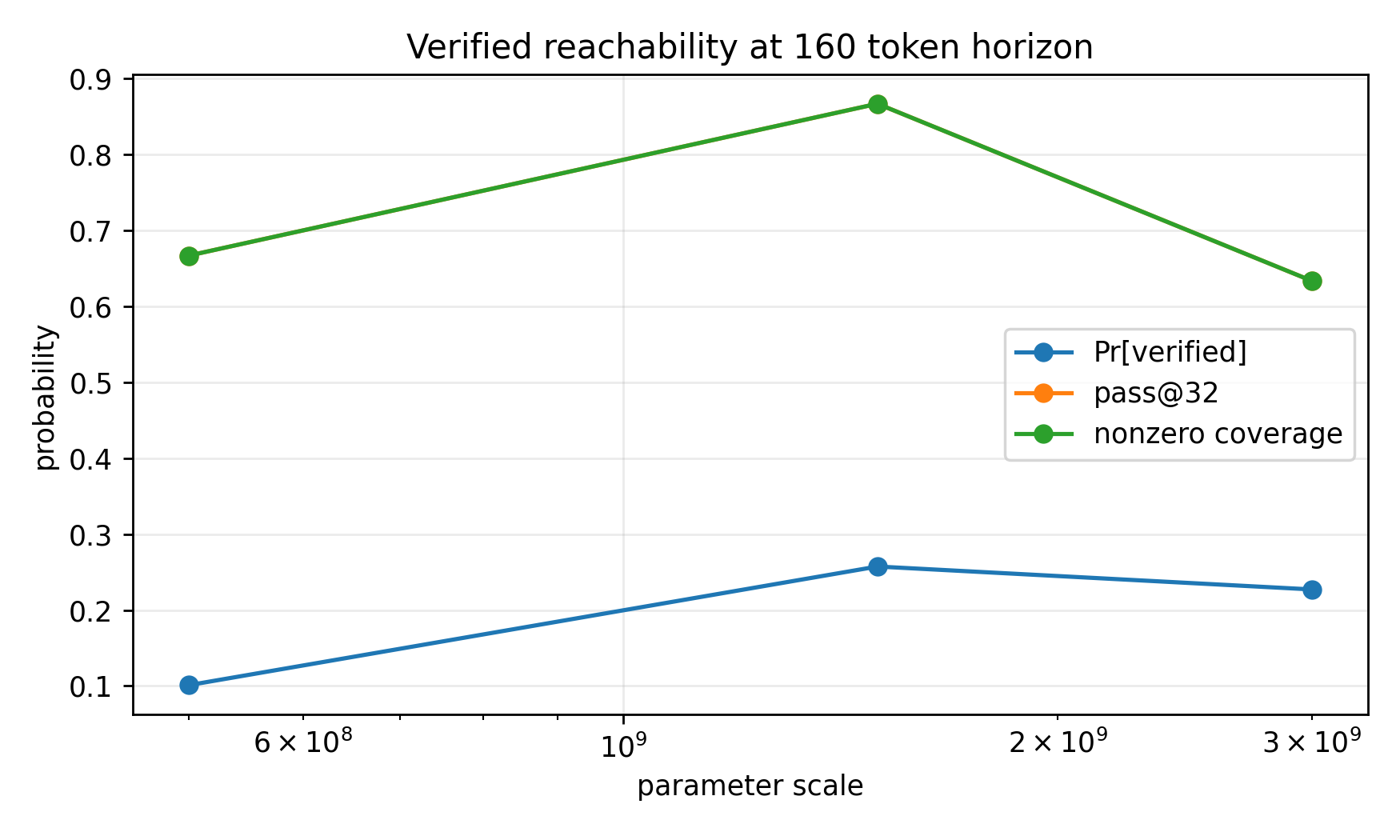}
\caption{Reachability.}
\end{subfigure}
\begin{subfigure}{0.32\linewidth}
\includegraphics[width=\linewidth]{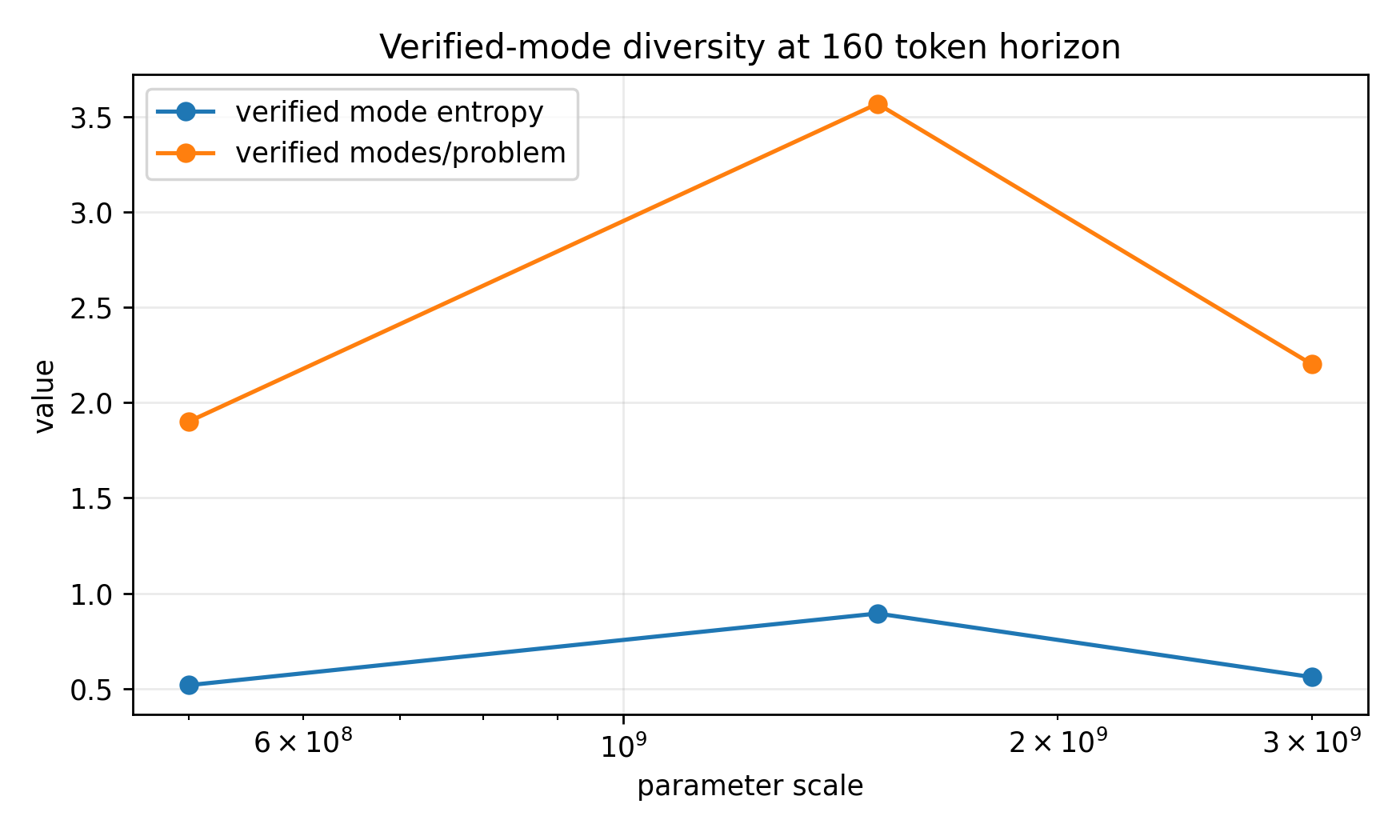}
\caption{Mode diversity.}
\end{subfigure}
\begin{subfigure}{0.32\linewidth}
\includegraphics[width=\linewidth]{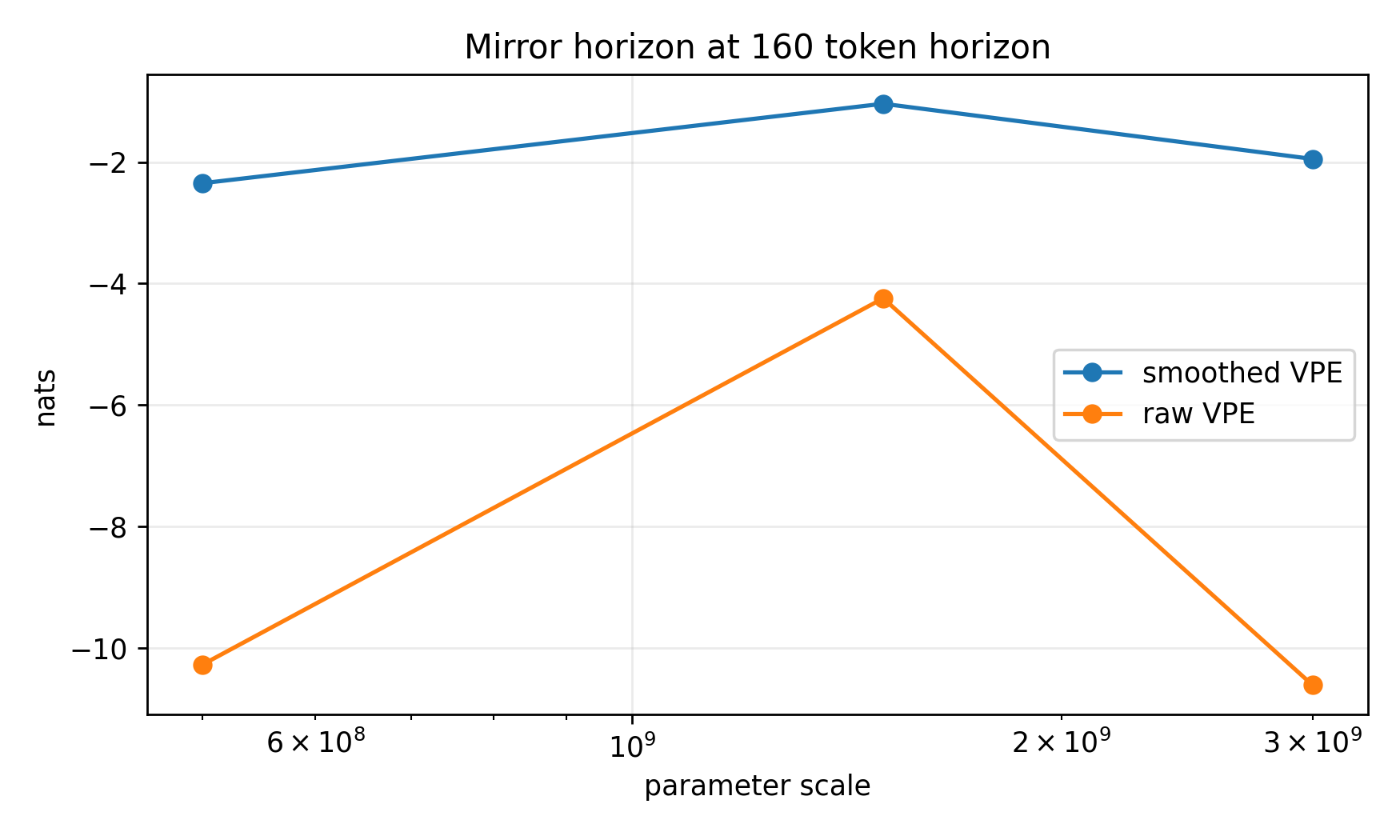}
\caption{VPE.}
\end{subfigure}
\caption{At 160 tokens, Qwen2.5-1.5B realizes the strongest accessible mirror horizon under the tested protocol. Qwen2.5-3B has higher average verified probability than 0.5B but weaker breadth and lower verified-mode diversity than 1.5B.}
\label{fig:scale160}
\end{figure}

\subsection{Per-problem decomposition}

\begin{figure}[t]
\centering
\includegraphics[width=0.95\linewidth]{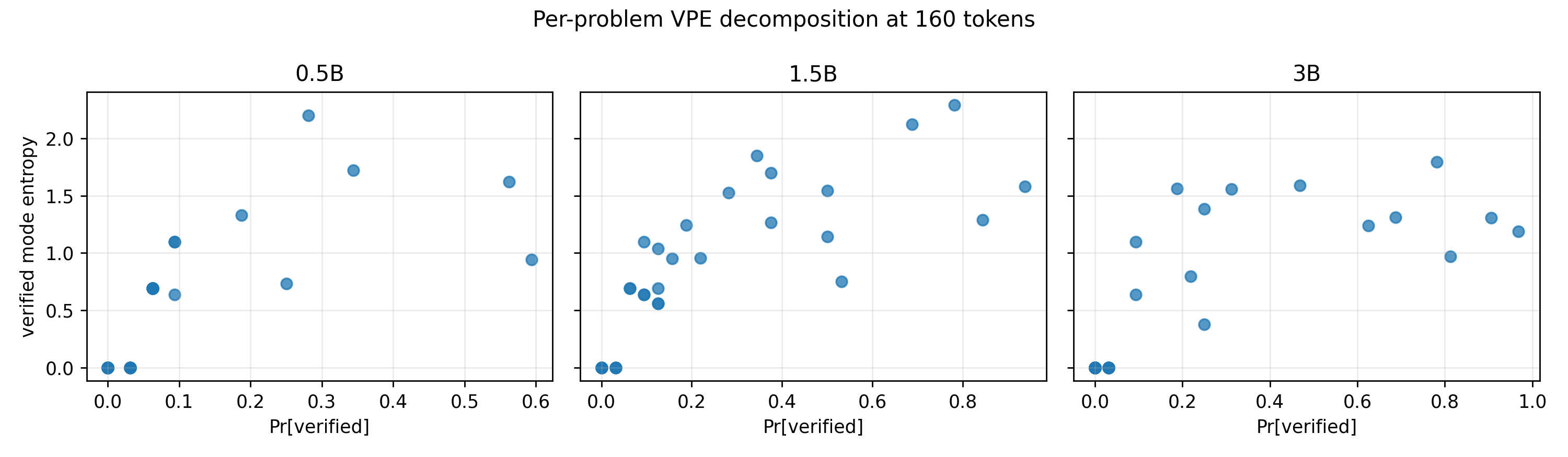}
\caption{Per-problem decomposition at 160 tokens. The x-axis is mean verified probability across 32 rollouts. The y-axis is verified-mode entropy. A strong mirror horizon requires both reachability and diversity.}
\label{fig:decomp}
\end{figure}

Figure~\ref{fig:decomp} plots each problem by verified probability and verified-mode entropy. Points at zero verified probability are unreachable under the rollout budget. Points with higher verified probability but low mode entropy indicate repeated success in a narrow mode. Points with both high verified probability and high entropy indicate robust, diverse verified continuation.

The 1.5B model has more points in the high-reachability, high-diversity region. The 0.5B model has several successes but many zero or low-reachability problems. The 3B model has some high-probability successes but fewer broad high-entropy regions than 1.5B. This supports the interpretation that 1.5B has the strongest accessible mirror horizon under this protocol.

\subsection{What the experiment displays for Mirror Theory}

The experiment directly instantiates the Mirror Theory chain:
\begin{equation}
    M \longmapsto \Prob^M_{B,T} \longmapsto \V(\gamma) \longmapsto \Mode(\gamma) \longmapsto \Hh_{B,T}(M).
\end{equation}
In the GSM8K setting, $M$ is the model and prompt-conditioned state, $B$ is the rollout count, $T$ is the token horizon, $\gamma$ is a sampled solution, $\V$ is numeric correctness, and $\Mode$ is the verified reasoning-mode map. The 96-to-160 token comparison then becomes a test of a simple prediction: increasing reflection horizon should reveal more viable continuations. It does. Both reachability and verified-mode diversity rise.

The result also supports the distinction between a static representation and a mirror. A one-shot accuracy score asks whether a model produces one correct answer. Mirror horizon asks what verified continuation structure unfolds under bounded reflection. This is why the 160-token result is more informative than the 96-token result: the mirror has more room to unfold, and the measured horizon expands.

\section{Axioms and empirical consequences}

The preceding definitions can be summarized as a small set of axioms. These are not meant to replace the path-space formalism; they state the modeling commitments that distinguish a mirror from a generic state.

\begin{axiom}[Indirectness]
The system acts through an internal mirror rather than directly on the world. In an LLM experiment, this means that the observed answer is mediated by the prompt-conditioned continuation law rather than by direct access to the mathematical ground truth.
\end{axiom}

\begin{axiom}[Finite reflection]
Mirror value is evaluated at finite budget and finite horizon. A mirror horizon is therefore not an intrinsic scalar attached to a model alone. It is indexed by the rollout count, token budget, verifier, mode map, and protocol.
\end{axiom}

\begin{axiom}[Viability]
A task supplies or induces a verifier. What survives reflection is not arbitrary diversity, but diversity filtered by task-specific invariants.
\end{axiom}

\begin{axiom}[Mode coarse-graining]
A task supplies or induces a semantic mode map. Token-level variety is not counted as mirror growth unless it corresponds to meaningful verified continuation modes.
\end{axiom}

\begin{axiom}[Constructive reflection]
A reflection protocol is useful to the extent that it increases viable continuation capacity. This can occur by increasing verified reachability, increasing verified-mode diversity, or both.
\end{axiom}

These axioms imply three empirical consequences. First, increasing reflection budget should often increase the measured horizon, but only when the additional budget is usable by the model and recognized by the verifier. Second, two systems with the same pass@k can differ in mirror horizon if one reaches more verified modes than the other. Third, the measured horizon may be nonmonotone in parameter count because parameter count is not the object being measured; accessible verified continuation capacity is.

\section{Detailed operationalization in language models}

This section spells out the mapping from the formal objects to our GSM8K experiment. The purpose is to make clear that the experiment is not an arbitrary benchmark add-on; it is an instantiation of the theory.

\paragraph{Mirror state.} We do not directly inspect hidden states in this first paper. We treat the prompt-conditioned model and its decoding protocol as inducing a mirror state. More precisely, for a fixed model $\theta$, prompt $q$, temperature, top-p, and token budget, the model induces a distribution over continuation paths. The mirror is this distribution viewed through the verifier and mode map.

\paragraph{Reflection budget.} The rollout count is 32 for each problem. The token horizon is either 96 or 160 maximum new tokens. The 96-to-160 comparison is therefore a direct reflection-budget intervention. It does not change model weights, task distribution, or sampling temperature; it changes how far the system may unfold a continuation.

\paragraph{Viability.} A rollout is viable if the extracted final numeric answer matches the GSM8K gold answer. This is intentionally stricter than human judgment. If a model reasons correctly but formats the answer in a way the extractor misses, the rollout is counted as nonviable. This is acceptable for a first operational measure because the verifier is explicit and reproducible. It also clarifies that mirror horizon is verifier-conditioned.

\paragraph{Modes.} A verified rollout is mapped to a coarse mode by length bin, arithmetic-operator signature, equation-count bin, and reasoning style. This mode map is intentionally simple. It provides a lower-resolution measurement of verified-mode diversity, not a final account of reasoning strategies.

\paragraph{Horizon.} For each problem and model, we compute the raw and smoothed horizon. The raw horizon is harsh: if no verified rollout appears among the 32 samples, the log term becomes very negative. The smoothed horizon substitutes a Jeffreys-style finite-sample correction for the viability probability. The paper reports both so the reader can distinguish theoretical harshness from finite-sample stability.

\paragraph{Why this is not merely pass@k.} pass@32 is a binary per-problem measure: at least one verified rollout exists or not. VPE is continuous and decomposed. It distinguishes a problem solved once from a problem solved many times, and it distinguishes repeated solutions in one mode from solutions distributed across several verified modes. Thus, VPE is not a replacement for pass@k; it explains what pass@k hides.

\section{Results in detail}

\subsection{Reflection budget as a direct test of mirror horizon}

The cleanest result is the 96-to-160 token comparison. It directly tests whether increasing the reflection horizon expands verified continuation capacity. The answer is yes for both models shared across the two runs.

For Qwen2.5-0.5B, verified probability increases from $0.057$ to $0.101$, pass@32 increases from $0.400$ to $0.667$, zero-verified fraction drops from $0.600$ to $0.333$, verified-mode entropy increases from $0.224$ to $0.518$, and smoothed VPE improves from $-3.249$ to $-2.347$. This is not merely an accuracy gain. The model reaches verified continuations on more problems and exhibits more verified-mode diversity among successful continuations.

For Qwen2.5-1.5B, the effect is stronger. Verified probability increases from $0.097$ to $0.257$, pass@32 from $0.567$ to $0.867$, zero-verified fraction drops from $0.433$ to $0.133$, verified-mode entropy rises from $0.507$ to $0.894$, and smoothed VPE improves from $-2.450$ to $-1.038$. The longer horizon therefore expands both breadth and depth: more problems become reachable, and successful problems support more verified modes.

This is the empirical result that most directly supports Mirror Theory. Reflection is not just repeated sampling. A longer finite horizon lets the model unfold more of its internal continuation structure. The verified part of that structure is what VPE measures.

\subsection{Model comparison at the 160-token horizon}

At 160 tokens, Qwen2.5-1.5B dominates all tested models. It has the highest verified probability, highest pass@32, lowest zero-verified fraction, highest verified-mode entropy, highest raw VPE, highest smoothed VPE, and highest average number of verified modes per problem. This makes it the strongest accessible mirror under the tested protocol.

The Qwen2.5-3B result is not monotone in parameter count. It produces more verified rollouts per problem than the 0.5B model on average, but it solves fewer problems at least once than 0.5B, and its verified-mode entropy is much lower than 1.5B. This indicates concentration: the 3B model can produce many correct continuations on some problems but fails to cover the task set as broadly as 1.5B under the same sampling rule and verifier.

This nonmonotonicity is a feature of the measurement, not a failure of the theory. Mirror Theory does not claim that larger parameter count is the primitive. It claims that the relevant object is the verified continuation structure actually reachable under bounded reflection. The 3B result is therefore informative: a larger model can have lower accessible horizon under a particular protocol.

\subsection{Breadth, depth, and zero-reachability}

The results naturally separate into breadth and depth. Breadth is captured by pass@32 and zero-verified fraction. Depth is captured by the number and entropy of verified modes among successful continuations. Qwen2.5-1.5B improves both. Qwen2.5-3B improves some depth-like quantities relative to 0.5B, such as average verified probability, but loses breadth. This breadth-depth separation is not visible in a single scalar benchmark score.

The zero-verified fraction is especially important. A problem with zero verified rollouts is not merely hard; under the current protocol, it has no observed viable continuation. Raw VPE penalizes such problems severely. This is appropriate under the theory, because a mirror cannot be credited for viable continuation modes it never reaches. Smoothed VPE softens the finite-sample penalty but preserves the same qualitative ranking.

\subsection{Why the theory belongs in the main paper}

The empirical result alone could be described as a new metric for sampled reasoning. That would be too small. The reason the result matters is that it operationalizes the earlier Mirror Theory thesis: mirrors are not static encodings, but structures that survive and unfold through reflection. The path-space definitions make this thesis measurable; the GSM8K experiment shows that the measurable object responds to reflection budget and reveals structure hidden by pass@k.

In this sense, the heavy theory is not decorative. Intuition, taste, geometry, and constructive compatibility explain why we should count verified continuation modes in the first place. Intuition supplies local constraints; taste selects what survives; geometry organizes which continuations are near, admissible, and stable; VPE measures the resulting finite-horizon capacity. Without this scaffold, the experiment would be an arbitrary metric. With it, the experiment becomes the first operational test of a theory of mirror horizon.

\section{Discussion}

\subsection{What this paper claims}

The claim is measure-level. Mirror Theory provides a path-space way to evaluate internal states by the viable continuations they support. VPE is one finite-sample estimator of that horizon. The experiments show that the measure is not empty: increasing reflection budget expands the measured horizon, and models with similar pass@k can differ in verified-mode diversity.

The paper does not claim a universal monotone law of parameter scaling. The 160-token result explicitly shows why that would be wrong. Qwen2.5-1.5B has the strongest horizon under the tested protocol, not Qwen2.5-3B. The correct conclusion is not that 3B is worse in general. It is that mirror horizon is accessible verified continuation capacity under a specified protocol. A larger model may contain more latent structure but fail to realize it under a particular prompt, sampling rule, verifier, or horizon.

\subsection{Why not just pass@k?}

If there is only one possible verified mode, mirror horizon reduces toward a pass@k-like reachability problem. But reasoning models often produce multiple qualitatively distinct verified solutions. Pass@k collapses those modes. VPE retains them. This matters because a system that can solve the same problem in multiple verified ways may be more robust to perturbations, transfer, or future reflection than a system that reaches only one brittle mode.

The current mode map is coarse, so this claim remains preliminary. But even a coarse map separates breadth from depth in the experiments. The next step is not to force a closed-form law; it is to test whether verified-mode coverage remains stable under richer mode maps and across tasks.

\subsection{Relation to taste and constructive compatibility}

The older theory of taste is not discarded. It is reinterpreted operationally. Taste is the selection pressure by which certain continuations survive. The verifier is an experimental proxy for invariant survival. The mode map is an experimental proxy for continuation geometry. Constructive compatibility is the increase in VPE caused by an update or protocol. In the present experiments, increasing the token horizon from 96 to 160 functions as a change in reflection budget. Its positive effect on VPE is evidence that a longer reflection window increases the mirror's viable continuation capacity.

\subsection{Relation to product and recommendation}

The constructive-compatibility note also motivated a product principle: recommend the most expansive possibility that the current mirror can still integrate. In the present paper, we do not evaluate recommendation systems. But the same structure appears. A possibility is valuable when it increases verified-mode coverage without collapsing viability. In reasoning, a longer reflection budget is such a possibility. In future work, prompts, tools, retrieved documents, or self-checking procedures can be evaluated by their VPE gain.

\section{What would count as verification of the revised definition}

The present experiments verify the first step of the revised definition: mirror horizon is measurable as finite-budget verified continuation capacity. A stronger verification program should test three additional properties.

\paragraph{Mode-map stability.} The qualitative ordering of models and budgets should not be an artifact of a single hand-written mode map. The heuristic mode map should be compared to TF-IDF clustering, embedding clustering, symbolic operation templates, and human-audited solution categories. The core question is whether the reachability component and the diversity component remain separately interpretable under different coarse-grainings.

\paragraph{Protocol sensitivity.} The theory predicts that the horizon is protocol-conditioned. Therefore, changing the prompt, token horizon, sampling temperature, or answer verifier should move the measured horizon in interpretable ways. A strict answer-format prompt may raise verified reachability; a higher temperature may raise diversity while lowering viability; a longer token horizon may reduce zero-reachability. These changes are not nuisance variables. They are part of the theory, because reflection is always bounded by a protocol.

\paragraph{Predictive value.} The most important future test is whether early-budget mirror horizon predicts later or out-of-distribution success better than pass@k alone. For example, one can compute VPE at $B=4$ or $B=8$ rollouts and ask whether it predicts pass@32, performance on held-out GSM8K problems, or transfer to SVAMP or MATH-mini. If verified-mode diversity predicts robustness beyond reachability, then the measure captures more than a reparameterization of pass@k.

These tests do not require a concrete closed-form scaling law. They remain at the measure level: a model and protocol induce a distribution over continuations; the verifier filters viable paths; the mode map coarse-grains the viable region; and the horizon measures how much verified structure is reachable. This is the right level for a first main-conference paper. The goal is not to assert a universal equation of scale, but to establish that the verified continuation law is a useful object of study.

\section{Limitations}

The experiments are intentionally small. They use 30 GSM8K problems, three models from one family, one temperature and top-p setting, one answer verifier, and one heuristic mode map. The mode map is coarse and may count surface structure rather than deep reasoning strategy. The numeric verifier may miss valid answers with unusual formatting. The sample size is sufficient for a pilot but not for a definitive scaling law.

The 96-token and 160-token runs are not meant to establish final model rankings. They establish that mirror horizon can be measured, that it changes systematically with reflection budget, and that it decomposes capability into interpretable components. A stronger version of the paper should add confidence intervals across problem subsets, more model families, more tasks, and multiple mode maps.

Finally, VPE depends on the verifier. This is a feature and a limitation. Mirror horizon is task-relative: a continuation is viable only relative to a criterion. This makes the measure operational, but it also means that bad verifiers give bad horizons.

\section{Future work}

Several next experiments are natural.

\paragraph{Larger coverage-curve studies.}
This paper now reports empirical coverage curves on the current 30-problem GSM8K pilot. A stronger version should repeat the same analysis with more problems, more rollout budgets, more model families, and multiple verifier-backed tasks. The goal is not to fit a closed-form curve, but to test whether verified-mode coverage remains stable and predictive across domains.

\paragraph{Mode-map robustness.}
The heuristic map should be compared against TF-IDF clusters, embedding clusters, symbolic operation templates, and human-audited reasoning modes. The theory becomes stronger if the qualitative findings survive changes in $\Mode$.

\paragraph{Protocol frontiers.}
The horizon is protocol-conditioned. Future experiments should vary max tokens, prompt format, temperature, self-consistency, tool use, and verifier strictness. The question is not only which model is larger, but which model-protocol pair realizes more verified continuation structure.

\paragraph{Task transfer.}
GSM8K is only the first test. The same measure should be evaluated on SVAMP, MultiArith, MATH subsets, HumanEval, chess puzzles, or formal proof tasks. Each provides a verifier and a potential mode map.

\paragraph{Post-training and compressed reasoning.}
Compressed reasoning work suggests that explicit, composed, and implicit reasoning traces expose different continuation structures \citep{matsutani2026zipping}. VPE can measure how post-training changes verified-mode coverage, not just final accuracy.

\section{Conclusion}

This paper restores Mirror Theory's formal scaffold and gives it a measurable empirical object. A mirror is not merely a representation. A mirror is an internal structure considered through the viable continuation law it induces. Viable path entropy measures the log-effective number of verified semantic continuation modes reachable under bounded reflection. Constructive compatibility is the increase in that horizon caused by integrating a possibility.

The GSM8K experiments provide the first operational test. Increasing the reflection horizon from 96 to 160 tokens expands verified reachability and verified-mode diversity. At 160 tokens, Qwen2.5-1.5B has the strongest accessible mirror horizon among the tested models. These results support the core claim: capability should be measured not only by one-shot correctness or pass@k, but by the verified continuation structure a system can sustain under bounded reflection.

\end{document}